\documentclass[journal,comsoc]{IEEEtran}
\usepackage[T1]{fontenc}
\usepackage{amsmath}
\usepackage[cmintegrals]{newtxmath}
\usepackage{bm}
\usepackage{graphicx}
\usepackage{multirow}
\usepackage{floatrow}
\usepackage{comment}
\usepackage{hyperref}
\usepackage{amssymb} 
\usepackage{color}

\usepackage{bm}
\usepackage{mathtools}
\usepackage{booktabs}
\usepackage{threeparttable}
\usepackage{tabularx}
\newcolumntype{Y}{>{\centering\arraybackslash}X}
\begin{document}
\title{Crowd Counting via Segmentation Guided Attention Networks and Curriculum Loss}
\author{Qian~Wang,~\IEEEmembership{Member,~IEEE,}
        Toby~P.~Breckon,~\IEEEmembership{Member,~IEEE,}
\thanks{Q. Wang is with the Department
of Computer Science, Durham University, United Kingdom, e-mail: qian.wang173@hotmail.com}
\thanks{TP. Breckon is with the Department of Computer Science and Department of Engineering, Durham University, United Kingdom, e-mail: toby.breckon@durham.ac.uk}
}

\markboth{Journal of xxx,~Vol.~14, No.~8, August~2020}%
{Wang \MakeLowercase{\textit{et al.}}: Crowd Counting via Segmentation Guided Attention Networks and Curriculum Loss}
\maketitle

\begin{abstract}
   Automatic crowd behaviour analysis is an important task for intelligent transportation systems to enable effective flow control and dynamic route planning for varying road participants. Crowd counting is one of the keys to automatic crowd behaviour analysis.
   Crowd counting using deep convolutional neural networks (CNN) has achieved encouraging progress in recent years. Researchers have devoted much effort to the design of variant CNN architectures and most of them are based on the pre-trained VGG16 model. Due to the insufficient expressive capacity, the backbone network of VGG16 is usually followed by another cumbersome network specially designed for good counting performance. Although VGG models have been outperformed by Inception models in image classification tasks, the existing crowd counting networks built with Inception modules still only have a small number of layers with basic types of Inception modules. To fill in this gap, in this paper, we firstly benchmark the baseline Inception-v3 model on commonly used crowd counting datasets and achieve surprisingly good performance comparable with or better than most existing crowd counting models. Subsequently, we push the boundary of this disruptive work further by proposing a Segmentation Guided Attention Network (SGANet) with Inception-v3 as the backbone and a novel curriculum loss for crowd counting. We conduct thorough experiments to compare the performance of our SGANet with prior arts and the proposed model can achieve state-of-the-art performance with MAE of 57.6, 6.3 and 87.6 on ShanghaiTechA, ShanghaiTechB and UCF\_QNRF, respectively.
\end{abstract}


\IEEEpeerreviewmaketitle

\section{Introduction}
\IEEEPARstart{A}{utomatic} crowd counting has attracted increasing attention in the research community since its valuable impacts in public surveillance and intelligent transportation systems \cite{zhou2018crowd,zhan2008crowd,ryan2015evaluation,sindagi2018survey,ding2020crowd}. Crowd behaviour can have a big effect in the efficiency of public transportation. Intelligent transportation systems deployed in a smart city should be able to capture real-time crowd behaviour information from public surveillance and dynamically adjust the planning for effective transportation. Accurate people and vehicle counting in varying conditions provide basic information for automatic crowd behaviour analysis. People and vehicle counting can be formulated in a unified object counting framework which aims to estimate the number of target objects in still images or video frames and has been applied in many real-world applications. For instance, there have been works focusing on automatic counting different objects including cells \cite{xie2018microscopy}, vehicles \cite{liang2015counting,moranduzzo2014automatic}, leaves \cite{giuffrida2016learning,aich2017leaf} and people \cite{sindagi2018survey}. 

In earlier years, crowd counting in images was implemented by detection \cite{zhao2003bayesian,dong2007fast,subburaman2012counting} or direct count regression \cite{kong2006viewpoint,siva2016real}. Counting by detection methods assume people signatures (i.e. the whole body or the head) in images are detectable and the count can be easily achieved from the detection results. This assumption, however, does not always hold in real scenarios, especially when the crowd is extremely dense. Counting by direct count regression aims to learn a regression model (e.g., support vector machine \cite{siva2016real} or neural networks \cite{kong2006viewpoint}) mapping the hand-crafted image features directly to the count of people in the image. Methods falling into this category only give the final counts hence lack of explainability and reliability.

Recently, crowd counting has been overwhelmingly dominated by density estimation based methods since the idea of density map was first proposed in \cite{lempitsky2010learning}. The use of deep Convolutional Neural Networks \cite{krizhevsky2012imagenet} to estimate the density map along with the availability of large-scale datasets \cite{zhang2016single,idrees2018composition} further improved the accuracy of crowd counting in more challenging real-world scenarios. Recent works in crowd counting have been focusing on the design of novel architectures of deep neural networks (e.g., multi-column CNN \cite{zhang2016single,sindagi2017generating} and attention mechanism \cite{liu2018decidenet,zhang2018attention}) for accurate density map estimation. The motivations of these designs are usually to improve the generalization to scale-variant crowd images. Among them, the \textit{Inception} module \cite{szegedy2016rethinking} has been employed and showed effectiveness in crowd counting \cite{cao2018scale,jiang2019crowd}, although only the basic \textit{Inception} modules are used and the networks are relative shallow compared with the state-of-the-art deep CNN models for image classification such as \textit{Inception-v3} \cite{szegedy2016rethinking} which uses heterogeneous \textit{Inception} modules to improve the expressive power of the network. Although VGG16, VGG19 and ResNet101 have been used as the backbone networks for crowd counting in \cite{guo2019dadnet,ma2019bayesian,wang2019learning}, to our best knowledge, the  \textit{Inception}  models have not been investigated.

In this paper, we make the first attempt to investigate the effectiveness of \textit{Inception-v3} model for crowd counting. We modify the original \textit{Inception-v3} to make it suitable for crowd density estimation. Without bells and whistles, the \textit{Inception-v3} model can achieve surprisingly good performance comparable with or even better than most existing crowd counting models on commonly used crowd counting datasets. Subsequently, we add a segmentation map guided attention layer to the \textit{Inception-v3} model to enhance the salient feature extraction for accurate density map estimation and propose a novel curriculum loss strategy to address the issues caused by extremely dense regions in crowd counting. As a result, the proposed SGANet with curriculum loss is able to achieve state-of-the-art performance for crowd counting with the embarrassingly simple design.
The contributions of this paper are summarized as follows:
\begin{itemize}
    \item [--] We make the first attempt to investigate the effectiveness of \textit{Inception-v3} in crowd counting and achieve disruptive results which are important for the research community.
    \item [--] We present a Segmentation Guided Attention Network (SGANet) with a novel curriculum loss function based on the \textit{Inception-v3} model for crowd counting.
    \item [--] Extensive evaluations are conducted on benchmark datasets and the results demonstrate the superior performance of SGANet and the effectiveness of curriculum loss in crowd counting.
\end{itemize}

The remainder of this paper is organized as follows.
Section \ref{sec:related} reviews related work of crowd counting and curriculum learning. In Section \ref{sec:sganet} we introduce our proposed segmentation guided attention networks for crowd counting with curriculum loss. Section \ref{sec:experiments} presents the experiments and results on several benchmark datasets and we conclude our work in Section \ref{sec:conclusion}.





\section{Related Work}\label{sec:related}
In this section, we first review related works on CNN based crowd counting and focus mainly on the diverse network architectures against which our proposed crowd counting model is compared. Subsequently, we introduce works related to curriculum learning and how they can potentially be used in the task of crowd counting.

\subsection{Crowd Counting Networks}\label{sec:networks}

Successful efforts have been devoted to the design of novel network architectures to improve the performance of crowd counting. Commonly used principles of network design for crowd counting include multi-column networks, rich feature fusion and attention mechanism.

Multi-column neural networks were employed to address the scale-variant issue in crowd counting \cite{zhang2016single,sam2017switching,cheng2019improving}. As one of the earliest CNN based models for crowd counting,
MCNN \cite{zhang2016single} consists of three branches aiming to handle crowds of different densities. Following this idea, Sam et al. \cite{sam2017switching} proposed SwithCNN which employs a classifier to explicitly select one of the three branches for a given input patch based on its level of crowd density. While these methods aim to use different kernel sizes in different branches to capture scale-variant information, Liu et al. \cite{liu2019crowd} proposed a model consisting of multiple branches of VGG16 networks with shared weights to process scaled input images respectively. Similarly, Ranjan et al. \cite{ranjan2018iterative} devised a two-column network which learns the low- and high-resolution density maps iteratively via two branches of CNN. 
The success of these specially designed network architectures has validated that multi-column CNN models are capable of capturing scale-variant features for crowd counting.

The second direction of network design is to pursue effective fusion of rich features from different layers \cite{sindagi2019ha,jiang2019crowd}. These attempts are based on the fact different layers have variant receptive fields hence capturing features of variant-scale information. Different feature fusion strategies including direct fusion \cite{sindagi2019ha}, top-down fusion \cite{sam2018top} and bidirectional fusion \cite{sindagi2019multi} have been employed in crowd counting. 

To take advantage of the two aforementioned ideas for crowd counting, one straightforward solution is to utilise the  \textit{Inception}  module \cite{szegedy2016rethinking} which was firstly proposed in \cite{szegedy2015going} and has evolved into a variety of more efficient forms to date. The \textit{Inception} modules have been employed in crowd counting models before in SANet \cite{cao2018scale} and the TEDNet \cite{jiang2019crowd}. Both of them use only the basic types of \textit{Inception} modules similar to those used in the first version of \textit{Inception} net (i.e. GoogLeNet \cite{szegedy2015going}). In our work, we aim to explore the more advanced \textit{Inception} modules in the framework of \textit{Inception-v3}.

The attention mechanism is another useful technique considered when designing network architectures for crowd counting \cite{liu2018decidenet,sindagi2019ha,guo2019dadnet,liu2019adcrowdnet}. Attention layers are usually combined with multi-column structures so that regions of different semantic information (e.g., background, sparse, dense, etc.) can be attended and processed by different branches respectively. Attention maps learned by these models have proved to be aware of semantic regions \cite{guo2019dadnet}, however, they cannot provide fine-grained scale awareness within the images. To address this issue explicitly, perspective maps have been employed to guide the accurate estimation of density maps \cite{zhang2015cross,onoro2016towards,shi2019revisiting}. In many scenarios where the perspective maps are not available, it is possible to estimate these perspective maps from the crowd images via a specially designed and trained network \cite{yan2019perspective}.

Alternatively, binary segmentation maps generated from point annotation \cite{zhao2019leveraging} are introduced as additional supervision for the training of crowd counting networks via multi-task learning \cite{zhao2019leveraging}. In our work, binary segmentation maps are treated as explicit attention maps guiding the learning of salient visual features for density map estimation. In this sense, our work is more related to \cite{sindagi2019inverse} and \cite{shi2019counting} in which the segmentation maps are also used as attention maps but in essentially different ways as explained in Section \ref{sec:architecture} and validated in Section \ref{sec:exp_sga}.


\subsection{Curriculum Learning} \label{sec:curriculum}
Curriculum learning is a strategy of model training (e.g., neural networks) in machine learning and was proposed by Bengio et al. \cite{bengio2009curriculum}. The idea of curriculum learning can date back to no later than 1993 when Elman \cite{elman1993learning} proved the benefit of training neural networks to learn a simple grammar by ``starting small". The strategy of curriculum learning is inspired by the way how humans learn knowledge from easy concepts to hard abstractions gradually. In the specific case of training a machine learning model, curriculum learning selects easy examples at the beginning of training and allows more difficult ones added to the training set gradually. A curriculum is usually defined as a ranking of training examples by some prior knowledge to determine the level of difficulty of a given example. Jiang et al. \cite{jiang2015self} extended curriculum learning to a so-called self-paced curriculum learning by integrated the ideas of original curriculum learning and self-paced learning \cite{kumar2010self} in a unified framework. 

In this work, we apply the strategy of curriculum learning in crowd counting to address the issue of large variance of the crowd density in the images. Curriculum learning has been employed for crowd counting in \cite{liu2019point} where the curriculum is designed on the \textit{image level}, i.e., a difficulty score is calculated for each training image. The training images are divided into multiple subsets based on their difficulty scores and the easiest subset is added into the training set first. By contrast, our curriculum learning strategy is characterized by a novel curriculum loss defined on the \textit{pixel level} as described in Section \ref{sec:cl}. We define that density map pixels of higher values than a \textit{threshold} have higher difficulty scores because these pixels are within regions of denser crowds. We use all training images throughout the training process but set the threshold to a low value at the beginning and increase it gradually so that the difficult pixels become easy ones and contribute more to the training. 
As a result, our curriculum learning strategy is simple to implement with zero extra cost and has been proved effective especially when there exist extremely dense crowd regions in the images.

\section{Segmentation Guided Attention Network}\label{sec:sganet}
\begin{figure*}
    \centering
    \includegraphics[width=\textwidth]{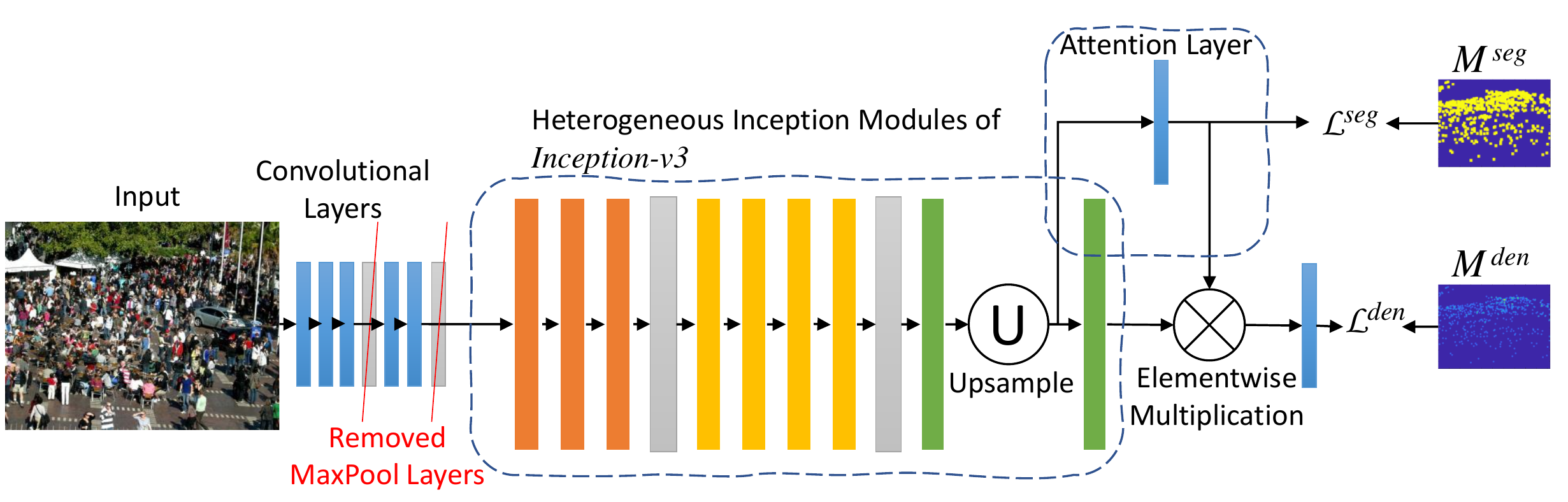}
    \caption{The framework of our proposed Segmentation Guided Attention Network (SGANet) which is adapted from \textit{Inception-v3} by: 1) removing the fully-connected layers; 2) removing two maxpooling layers to reserve high spatial resolution feature maps; 3) adding an upsampling layer before the last \textit{Inception} module; 4) adding an attention layer whose output is applied to the feature maps generated by the last \textit{Inception} module; 5) adding a convolutional layer for density map estimation.}
    \label{fig:framework}
\end{figure*}

Crowding counting is formulated as a density map regression problem in this study. Given a crowd image $I$, we aim to learn a Fully Convolutional Network (FCN) denoted as $\mathcal{F}$ so that the corresponding density map $M^{den}$ can be estimated by:
\begin{equation}
    \label{eq:fcn}
    \hat{\bm{M}}^{den} = \mathcal{F}(I;\bm{\Theta}).
\end{equation}
where $\bm{\Theta}$ is a collection of parameters of the FCN.

As shown in Figure \ref{fig:framework}, our proposed network is adapted from the famous \textit{Inception-v3} originally designed for image classification by Google Research \cite{szegedy2016rethinking}. We first modify \textit{Inception-v3} to an FCN so that it can process images of arbitrary sizes and generates the estimated density maps $M^{den}$ as the outputs. An attention layer is added to the network to filter out features within the background region and concentrate on the foreground features for accurate density map estimation. Since the attention maps generated by this attention layer aim to discriminate the regions of background and foreground of the feature maps, we use a ground truth segmentation map,  which can be easily derived from point annotations, as extra guidance for the training of the attention layer. As a result, the learned attention maps are forced to be similar to the segmentation maps during training. 

We also investigate the use of curriculum loss in the training of crowd counting networks. Specifically, we define a curriculum based on the pixel-wise difficulty level so that the network starts training by focusing more on the ``easy" regions (sparse) within the density maps and down-weighting the ``hard" pixels (dense). During training, the ``hard" pixels are gradually exposed to the model and finally, the learned model can perform well for all situations.

\subsection{Density and Segmentation Maps} \label{sec:segMap}
In this study, we use simple ways to generate density and segmentation maps from the point annotations although more complicated ones \cite{shi2019counting} might benefit the performance. For density maps $\bm{M}^{den} \in \mathbb{R^+}^{H\times W}$, where $H$ and $W$ are the height and width of the image, we follow \cite{zhang2016single} using a Gaussian kernel $G_\sigma \in \mathbb{R^+}^{15\times 15}$ with fixed $\sigma=4$:
\begin{equation}\label{eq:denmap}
\bm{M}^{den}(\bm{x}) = \sum_{i=1}^N \delta(\bm{x}-\bm{x}_i)*G_{\sigma}(\bm{x}).
\end{equation}

For segmentation maps $\bm{M}^{seg} \in \{0,1\}^{H\times W}$, we use a similar method:
\begin{equation}
    \label{eq:segmap}
    \bm{M}^{seg} (\bm{x}) = \sum_{i=1}^N \delta(\bm{x}-\bm{x}_i)*J_n(\bm{x}), 
\end{equation}
where $J_n(\bm{x})$ is an all-one matrix of size $n\times n$ centred at the position $\bm{x}$. As a result, ones and zeros in the matrix $\bm{M}^{seg}$ denote the pixels belong to the foreground and background regions, respectively. We empirically set $n=25$ across all our experiments to ensure that a specific head within an image is characterized by more pixels in the segmentation map than in the density map to avoid losing useful contextual information.
\subsection{Network Configuration} \label{sec:architecture}
Instead of designing a novel network from scratch, we exploit the state-of-the-art CNN model for image classification \textit{Inception-v3} in our study. 
To apply the original \textit{Inception-v3} network in crowd counting, some favourable modifications have been made. Firstly, we remove the final fully-connected layers and reserve all the convolutional layers. The input size of the original \textit{Inception-v3} network is $299\times299$ and the output size of the final convolutional layer is $8\times8$. That is to say, feature maps generated by the last convolutional layer have approximately $\frac{1}{2^5}$ spatial resolutions of the input image. This is achieved by the first convolutional layers (stride of 2), two max-pooling layers (stride of 2) and two \textit{Inception} modules in which max-pooling (stride of 2) is employed. To ensure the spatial resolution of the output density map which is important in crowd counting, we remove the first two max-pooling layers from the original network and add one upsample layer before the final \textit{Inception} module. As a result, the output of the modified network has exactly $\frac{1}{4}$ spatial resolution of the input image when the input size is $2^n$ (e.g., $128\times128$ in our case). Such modification does not change the number of parameters of the network hence the pre-trained weights can be directly loaded and used. However, since the spatial resolutions of intermediate feature maps have been increased, the number of operations is also increased. This modified model will also denoted as \textit{Inception-v3} without introducing ambiguity and used as a baseline method in our experiments.

Distinct from existing works using the segmentation map in the framework of multi-task learning \cite{zhao2019leveraging} to extract more salient features for density map estimation, we claim that the segmentation map can be used as an ideal attention map to emphasize the contributions of features within the foreground regions to the density map estimation whilst compressing the effects of features within the background regions. To this end, we add an attention layer to estimate the attention map. The attention layer is a convolutional layer followed by a sigmoid layer which restricts the output values in the range of 0--1. The attention layer takes the feature maps generated by the second last \textit{Inception} module as input and outputs a one-channel attention map of the same spatial resolution as the input. Subsequently, the attention map estimated by the attention layer is applied to the feature map generated by the last \textit{Inception} module by an element-wise product with each channel of the feature map.
\begin{equation}
    \label{eq:attention}
    F^l = F^{l-1} \odot M_{att}
\end{equation}

The attention layer designed in our framework is similar to that in \cite{sindagi2019inverse,shi2019counting}. However, a so-called inverse attention map is estimated in \cite{sindagi2019inverse} while our attention layer generates an attention map directly applied to the feature map. Also, the foreground regions in the ground truth segmentation map in \cite{sindagi2019inverse} are derived by thresholding the density map hence both maps have the same positive fields for each head while ours are different (c.f. Eq.(\ref{eq:denmap}-\ref{eq:segmap})). In \cite{shi2019counting}, the attention layer takes the feature map as input to estimate an attention map which again is applied to the same feature map. This may limit the capacity of the model since it is forced to learn two different maps from the same feature map via two convolutional layers which have limited parameters. In contrast, as mentioned above, the input of our attention layer is the feature map from the previous layer which has higher spatial resolutions and is different from the one the generated attention map will be applied to. These favourable distinctions collectively benefit the estimation of the density map and will be empirically evaluated in our experiments.

\subsection{Loss Function} \label{sec:loss}
We first describe the loss function used to train the SGANet without curriculum loss in this section and describe the curriculum loss in the following section. The loss function consists of two components. The first one is an L2 loss applied to the estimation of the density map and is denoted as $\mathcal{L}^{den}$. The density map loss can be calculated as follows:
\begin{equation}
    \label{eq:denloss}
    \mathcal{L}^{den}(\bm{\Theta}) = \frac{1}{2N} \sum_{i=1}^N ||\hat{\bm{M}}_i^{den} - \bm{M}^{den}_i||_F^2.
\end{equation}
The second component of the loss function is the segmentation map loss $\mathcal{L}^{seg}$ which is defined as the cross-entropy loss:
\begin{equation}
    \label{eq:segloss}
    \begin{multlined}
    \mathcal{L}^{seg}(\bm{\Theta}) = -\frac{1}{N} \sum_{i=1}^N ||\bm{M}_i^{seg} \odot log(\hat{\bm{M}}_i^{seg}) \\
    + (1-\bm{M}_i^{seg})\odot log(1-\hat{\bm{M}}_i^{seg})||_1.
    \end{multlined}
\end{equation}
where $||\cdot||_1$ denotes the elementwise matrix norm, i.e., the sum of all elements in a matrix, and $\odot$ denotes elementwise multiplication of two matrices with the same size.
These two components are combined during network training and the compositional loss function is:
\begin{equation}
    \label{eq:loss}
    \mathcal{L}(\bm{\Theta}) = \mathcal{L}^{den}(\bm{\Theta}) + \lambda \mathcal{L}^{seg}(\bm{\Theta}) 
\end{equation}
where $\lambda$ is a hyper-parameter which ensures the two components to have comparable values and is set 20 across our experiments.

\subsection{Curriculum Loss} \label{sec:cl}
To benefit from the strategy of curriculum learning, we present a novel curriculum loss function in this section to replace the traditional density map loss function defined in Eq. (\ref{eq:denloss}). The curriculum loss function is designed to be aware of the pixel-wise difficulty level when computing the density map loss. Based on the fact that dense crowds are generally more difficult to count than sparse ones, we design a curriculum where pixels of higher values than a dynamic threshold in the density map are defined as difficult pixels. We set the dynamic threshold and assign variant weights to different pixels of the density map when calculating the density map loss. Specifically, we define a weight matrix $\bm{W}$ as follows:
\begin{equation}
    \label{eq:weight}
    \bm{W} = \frac{T(e)}{max\{\bm{M}^{den}-T(e),0\}+T(e)}.
\end{equation}
The weight matrix $\bm{W}$ has the same size as the density map matrix $\bm{M}^{den}$ used in Eq.(\ref{eq:denloss}) and the pixel-wise weights are determined by the dynamic threshold $T(e)$ and the pixel values in the density map. 
If the pixel value of the density map is higher than the threshold, this pixel is treated as a difficult one and the corresponding weight is set less than one, otherwise the weight is equal to one. The higher the pixel values are, the smaller the weights will be. As a result, the training will focus more on the pixels of lower density value than $T(e)$.

The dynamic threshold $T(e)$ is defined as a function of the training epoch index in the form of:
\begin{equation}
    \label{eq:threshold}
    T(e) = k e + b
\end{equation}
where $k$ and $b$ can be determined based on the prior knowledge of the pixel values in the ground truth density map. The value of $b$ is the initial threshold which should be equivalent to the maximum density value in the region characterizing a single head. The value of $k$ controls the speed of increasing the difficulty which can also be easily derived from the learning curve when the curriculum learning strategy is not used.

Finally, the curriculum loss function for density map can be derived by modifying Eq.(\ref{eq:denloss}) as:
\begin{equation}
    \label{eq:weightdenloss}
    \mathcal{L}^{den}(\bm{\Theta}) = \frac{1}{2N} \sum_{i=1}^N ||\bm{W}(e)\odot (\hat{\bm{M}}_i^{den} - \bm{M}^{den}_i)||_F^2.
\end{equation}
where $\bm{W}(e)$ is also a function with respect to the training epoch index $e$.

\section{Experiments}\label{sec:experiments}
Extensive experiments have been conducted on benchmark datasets to evaluate the performance of SGANet and the effectiveness of curriculum loss in crowd counting. We will briefly describe the datasets and evaluation metrics used in our experiments, details of experimental protocols and network training. Experimental results are compared with state-of-the-art methods and analysed. We also present an ablation study to investigate the contributions of different components to the performance of the proposed framework.

\subsection{Datasets}\label{sec:dataset}
\quad \textbf{ShanghaiTech} dataset was collected and published by Zhang et al. \cite{zhang2016single} consisting of two parts. Part A consists of 300 and 182 images of different resolutions for training and testing respectively. The minimum and maximum counts are 33 and 3139 respectively, and the average count is 501.4. Part B consists of 400 and 316 images of a unique resolution (768$\times$1024) for training and testing respectively. Compared with part A, the numbers of people in these images are much smaller with the minimum and maximum counts of 9 and 578 respectively, and the average count is 123.6.

\textbf{UCF\_QNRF} dataset \cite{idrees2018composition} contains 1,535 high-quality images, among which 1201 images are used for training and 334 images for testing. The minimum and maximum counts are 49 and 12,865 respectively, and the average count is 815.

\textbf{UCF\_CC\_50} dataset \cite{idrees2013multi} contains 50 images with the minimum and maximum counts of 94 and 4,534 respectively. It is a challenging dataset due to the limited number of images. Following the suggestion in \cite{idrees2013multi} and many other works, we use 5-fold cross-validation in our experiments. 

\subsection{Evaluation Metrics} \label{sec:evaluation}
We follow the previous works using two metrics, i.e., the mean absolute error (MAE) and the root mean squared error (RMSE), to evaluate the performance of different models in our experiments. The two metrics can be calculated as follows:
\begin{equation}
    \label{eq:mae}
    MAE = \frac{1}{N_{test}} \sum_{i=1}^{N_{test}} |y_i-\hat{y}_i|
\end{equation}
\begin{equation}
    \label{eq:rmse}
    RMSE = \sqrt{\frac{1}{N_{test}}\sum_{i=1}^{N_{test}} (y_i-\hat{y}_i)^2}
\end{equation}
where $y_i$ and $\hat{y}_i$ are the ground truth and predicted count for $i$-th test image respectively, $N_{test}$ is the number of test images.

\subsection{Network Training}
SGANet is implemented in PyTorch \cite{paszke2017automatic} and the source code is publicly available \footnote{https://github.com/hellowangqian/sganet-crowd-counting}. The ``Adam" optimizer \cite{kingma2014adam} is employed for training. The initial learning rate is set to 1e-4 and reduced by a factor of 0.5 after every 50 epochs. The total number of training epochs is set 500 since the model can always converge much earlier than that. The network is trained with image patches with a size of $128\times 128$ randomly cropped from the training images. Instead of preparing the patches in advance, we do the random patch cropping on-the-fly during training. Specifically, we randomly select 8 images from the training set and 4 patches are randomly cropped from each selected image. This leads to a batch of 32 training patches in each iteration of training. The training patches generated in this way can be more diverse and help to alleviate the potential over-fitting problem. Since the output of SGANet has the size of $32\times 32$ (i.e. 1/4 of the input size), we use sum-pooling to adapt the ground truth density and segmentation map so that they have the same size of $32\times 32$ as the output. The training patches, as well as their corresponding density and segmentation maps, are horizontally flipped with a probability of 0.5 for data augmentation which has been shown beneficial in many works \cite{guo2019dadnet,zhang2019relational}. For testing, we feed the whole image into the network and obtain the density map from which the predicted count can be computed. For the UCF\_QNRF dataset, to save the memory usage during testing, we also resize the images from both training and test sets so that all images are limited to have resolutions no higher than 2048 whilst the original aspect ratios are kept, if not specified otherwise.


\begin{table*}[t]
\centering
    {
        \centering
        \caption[]{Comparison results with state-of-the-art models for crowd counting ( -- denotes the results are not available; CL denotes curriculum loss).
        }
        \label{table:comparative}
        \begin{tabularx}{0.72\textwidth}{l*{8}{|Y} }
        \hline
        \multirow{2}{*}{Model}  &  \multicolumn{2}{c|}{ShTechA} & \multicolumn{2}{c|}{ShTechB} &\multicolumn{2}{c|}{UCF-QNRF} &\multicolumn{2}{c}{UCF-CC-50} \\ \cline{2-9}
        & MAE & RMSE& MAE & RMSE & MAE & RMSE& MAE & RMSE\\ \hline
        MCNN \cite{zhang2016single} & 110.2 & 173.2 & 26.4 & 41.3 & -- & -- & 377.6 & 509.1 \\
        CSRNet \cite{li2018csrnet}  & 68.2 & 115.0 & 10.6 & 16.0 & -- & -- & 266.1 & 397.5 \\
        SANet \cite{cao2018scale}  & 67.0 & 104.5 & 8.4 & 13.6 & -- & -- & 258.4 & 334.9 \\
        DADNet \cite{guo2019dadnet}  & 64.2 & 99.9 & 8.8 & 13.5 & 113.2 & 189.4 & 285.5 & 389.7\\
        CANNet \cite{liu2019context}  & 62.3 & 100.0 & 7.8 & 12.2 & 107 & 183 & \bf 212.2 & \bf 243.7\\
        TEDNet \cite{jiang2019crowd} & 64.2 & 109.1 & 8.2 & 12.8 & 113 & 188 & 249.4 & 354.5 \\
        Wan et al. \cite{wan2019adaptive}  & 64.7 & \underline{\it97.1} & 8.1 & 13.6 & 101 & 176 & -- & -- \\
        RANet\cite{zhang2019relational} & 59.4 & 102.0 & 7.9 & 12.9 & 111 & 190 & 239.8 & 319.4\\
        ANF \cite{zhang2019attentional}  & 63.9 & 99.4 & 8.3 & 13.2 & 110 & 174 & 250.2 & 340.0\\
        SPANet \cite{cheng2019learning} & 59.4 & \textbf{92.5} & 6.5 & \underline{\it9.9} & -- & -- & 232.6 & 311.7 \\
        
        \hline
        {Inception-v3}  & 60.1 & 105.0 & \underline{\it 6.4} & \textbf{9.8} & 95.6 & 165.4 & 236.0 & 304.9 \\
        SGANet  & \underline{\it58.0} & 100.4 & \textbf{6.3} & 10.6 & \underline{\it 89.1} & \textbf{150.6} &224.6 & 314.6 \\
        SGANet + CL  & \textbf{57.6}  & 101.1 &6.6 & 10.2  & \textbf{87.6} & \underline{\it 152.5} & \underline{\it 221.9} & \underline{\it 289.8} \\
        \hline

    \end{tabularx}
}
\end{table*}
\subsection{Comparative Study} \label{sec:comparative}
We select both classic and state-of-the-art models for the comparison, including \textbf{MCNN} \cite{zhang2016single} which is a three-column CNN, \textbf{CSRNet} \cite{li2018csrnet} which uses VGG16 as the front-end and dilated convolutional layers as the back-end, \textbf{SANet} \cite{cao2018scale} which employs the basic \textit{Inception} modules but has a relatively shallow depth, \textbf{DADNet} \cite{guo2019dadnet} which employs the ideas of dilated convolution, attention map and deformable convolution in the framework, \textbf{CANNet} \cite{liu2019context} which captures context-aware feature by multiple branches, \textbf{TEDNet} \cite{jiang2019crowd} which also uses \textit{Inception}-style modules, \textbf{RANet} \cite{zhang2019relational} which uses an iterative distillation algorithm, \textbf{ANF} \cite{zhang2019attentional} which uses conditional random fields (CRFs) to aggregate multi-scale features, and \textbf{SPANet} \cite{cheng2019learning}.

The experimental results are listed in Table \ref{table:comparative} where the best result in each column is highlighted in \textbf{bold} and the second best in underscored \textit{italic}. From Table \ref{table:comparative}, we can see our modified \textit{Inception-v3} can achieve very competitive performance on all four datasets. Especially on ShanghaiTech part B, it achieves the second best MAE of 6.4 and the best RMSE of 9.8. On the UCF$\_$QNRF dataset, \textit{Inception-v3} also achieves significantly better results than 
most existing models including TEDNet (MAE: 95.6 vs 113 and MSE: 165.4 vs 188) which also employs the \textit{Inception} modules. These results demonstrate the superiority of heterogeneous \textit{Inception} modules in classification problems can be transferred to the task of crowd counting hence different \textit{Inception} modules deserve more attention when designing a novel CNN architecture for crowd counting as well as other tasks suffering from the issue of scale variance. On the other hand, the disruptive performance of \textit{Inception-v3} in crowd counting provides more insight for the research community regarding the selection of backbone models when designing novel network architectures for crowd counting.

By adding the segmentation guided attention layer, our SGANet can achieve better performance on all datasets in terms of MAE (i.e. 58.0 vs 60.1 for ShanghaiTech part A, 89.1 vs 95.6 for UCF-QNRF and 224.6 vs 236.0 for UCF-CC-50), although the improvement on ShanghaiTech part B dataset is very marginal (i.e. 6.3 vs 6.4). Regarding RMSE, SGANet achieves better performance on ShanghaiTech part A (i.e. 100.4 vs 105.0) and UCF$\_$QNRF (i.e. 150.6 vs 165.4) but worse results on the other two datasets (i.e. 10.6 vs 9.8 for ShanghaiTech part B and 314.6 vs 304.9 for UCF\_CC\_50). Overall, our proposed SGANet with the combination of \textit{Inception-v3} and a segmentation guided attention layer can achieve state-of-the-art performance on several benchmark datasets.

The use of curriculum loss (SGANet+CL) further improves the performance of SGANet on three out of four datasets and these three datasets (i.e. ShanghaiTech part A, UCF\_QNRF and UCF\_CC\_50) consist of crowds with significant density variations. On the ShanghaiTech part B dataset, the use of curriculum loss does not improve the performance because the images from this dataset contain crowds with a relatively small variance of head scales. These results provide evidence that the issue of large scale variance can be further alleviated by the use of our proposed curriculum loss.  We will provide more evidence for the effectiveness of curriculum loss in the following ablation study.

\begin{table}[htbp]
     \centering
    {
        \caption[]{The effect of curriculum learning in different models on ShanghaiTech part A (the symbol $\downarrow$ means the error decreases with the use of curriculum loss).}
    \label{table:cl}
    \resizebox{0.85\columnwidth}{!}{%
    \begin{tabular}{l|l|l|l|l}
        \hline
        \multirow{2}{*}{Model} & \multicolumn{2}{c|}{Without CL} & \multicolumn{2}{c}{With CL}\\ \cline{2-5}
        & MAE & RMSE & MAE & RMSE \\ \hline
        MCNN  &91.8 & 144.9& 89.1 $\downarrow$  & 142.3 $\downarrow$ \\
        CSRNet & 67.2 & 110.5 & 66.7 $\downarrow$ & 113.7 \\
        SANet  & 64.0 & 103.4& 62.1 $\downarrow$  & 100.3 $\downarrow$ \\
        CANNet  & 65.6 & 106.7& 63.9 $\downarrow$  & 103.9 $\downarrow$ \\
        DADNet  & 63.7 & 107.4 & 64.2 & 102.1 $\downarrow$\\
        Inception-v3 & 60.1 & 105.0 &  58.2 $\downarrow$ & 97.9 $\downarrow$ \\
        \hline
    \end{tabular}%
	}
}
\end{table}

\begin{table*}[htbp]
     \centering
    {
        \caption[]{The effect of image resolutions in the performance of SGANet on UCF$\_$QNRF dataset.}
    \label{table:imgresize}
    \resizebox{0.6\columnwidth}{!}{%
    \begin{tabular}{l|c|c|c|c|c|c}
        \hline
        \multirow{2}{*}{Model} & \multicolumn{2}{c|}{UCF$\_$QNRF$\_$512} & \multicolumn{2}{c|}{UCF$\_$QNRF$\_$1024} & \multicolumn{2}{c}{UCF$\_$QNRF$\_$2048}\\ \cline{2-7}
        & \: MAE \: & RMSE & \: MAE \:& RMSE & \: MAE \: & RMSE \\ \hline
        SGANet  &126.1 & 236.1 & 102.5 & 178.4 & 89.1  & 150.6 \\
        SGANet + CL & 118.5 & 217.2 & 97.5 & 169.2 & 87.6 & 152.5 \\
        \hline
        Performance gain & 7.6 & 18.9 & 5.0 & 9.2 & 1.5 & -1.9 \\
        \hline
    \end{tabular}%
	}
}
\end{table*}

\subsection{Results on Curriculum Loss}\label{sec:result_cl}
The use of curriculum loss has shown a positive effect when applied to SGANet for crowd counting (Table \ref{table:comparative}). In this section, we attempt to explore the effectiveness of curriculum learning in the training of other crowd counting networks. To this end, we consider ``MCNN", ``CSRNet", ``SANet", ``CANNet", ``DADNet" and our modified ``Inception-v3" and use the curriculum loss when training these networks on ShanghaiTech part A. Firstly, we try to reproduce the results of these crowd counting models using conventional density map loss under our training protocols to remove the effects of various factors such as the ways of density map generation, patch cropping, data augmentation and so on for a fair comparison and focus on the effect of curriculum loss. It is noteworthy that the generated density maps have different sizes for these models (e.g., the size ratio between input and output is 1 for ``SANet", 2 for ``DADNet", 4 four ``MCNN" and ``Inception-v3", 8 for ``CSRNet" and ``CANNet"). The ground truth density maps need to be resized by sum pooling to have the same size as the corresponding outputs. As a result, the pixel values of the ground truth density maps for different models will have different distributions. This leads to model-specific curriculum designs (i.e. the parameter values in Eq.(\ref{eq:threshold})). 
Specifically, we set $b$ as the maximum value in the Gaussian kernel matrix $G_{\sigma}$ used for density map generation (c.f. Eq. (\ref{eq:denmap})) so that the sparse crowd regions without annotation overlapping will not be affected throughout the training process. The value of $k$ in Eq. (\ref{eq:threshold}) is determined by the number of epochs so that all the crowd regions will contribute to the loss equally before training is finished. 
In our experiments, we set $k=1e-3$ and $b=0.1$ for SGANet. Experimental results are shown in Table \ref{table:cl}. The use of curriculum loss improves the performance of most models. Specifically, the MAE decrease for all models except ``DADNet" and the RMSE decrease for all models except ``CSRNet". These experimental results demonstrate the curriculum loss is useful not only for our SGANet but also many other crowd counting models.

To evaluate the effect of crowd density in the performance of SGANet and the curriculum loss, an additional experiment is conducted on the UCF$\_$QNRF dataset. As mentioned above, we have changed the image resolutions in this dataset to be no higher than 2048 for computation efficiency. In this experiment, we create two more datasets by setting the image resolution thresholds as 1024 and 512 respectively. As a result, the images in the UCF$\_$QNRF$\_$512 dataset will have higher crowd density than those in the UCF$\_$QNRF$\_$1024 dataset which again consists of denser crowds than the UCF$\_$QNRF$\_$2048 dataset. We use SGANet on these three datasets and the experimental results are shown in Table \ref{table:imgresize}. It is obvious the image resolutions make a significant different in the performance and the models perform the best on the UCF$\_$QNRF$\_$2048 dataset whose image resolutions are higher hence have less crowded images. By comparing the performance of SGANet without and with curriculum loss, the use of curriculum loss leads to better results on all three datasets in terms of both MAE and RMSE except that in the last column of Table \ref{table:imgresize}. The performance gains achieved by the use of curriculum loss are also related to the image resolutions or the crowd densities in the datasets. 
Specifically, the MAE and RMSE are reduced by 7.6 and 18.9 respectively on UCF\_QNRF\_512, 5.0 and 9.2 on UCF\_QNRF\_1024, 1.5 and -1.9 on UCF\_QNRF\_2048. These results provide more evidence that the use of curriculum loss is more effective when the crowds are denser in the images.

In summary, the experimental results in Tables \ref{table:comparative}--\ref{table:imgresize} provide sufficient evidence that the use of curriculum learning can benefit the training of crowd counting models in most cases especially when the head scales vary a lot in the crowd images.

\subsection{Results on Segmentation Guided Attention}\label{sec:exp_sga}

From Table \ref{table:comparative} we can see the performance enhancement contributed by the segmentation guided attention layer by comparing the performance between \textit{Inception-v3} and SGANet. To validate the superiority of our segmentation guided attention layer to other similar designs \cite{shi2019counting}, we conduct an experiment on ShanghaiTech part A and UCF$\_$QNRF. In this experiment, we follow \cite{shi2019counting} and modify the SGANet by feeding the feature maps of the last \textit{Inception} module into the attention layer and keeping the rest unchanged. The experimental results are shown in Table \ref{table:exp_sga} from which we conclude the way segmentation maps are used in our SGANet outperforms that in \cite{shi2019counting}.

\begin{table}[htbp]
     \centering
    {
        \caption[]{Results of different approaches to segmentation map supervision.}
    \label{table:exp_sga}
    \resizebox{0.95\columnwidth}{!}{%
    \begin{tabular}{l|c|c|c|c}
        \hline
        \multirow{2}{*}{Model} & \multicolumn{2}{c|}{ShTechA } & \multicolumn{2}{c}{UCF$\_$QNRF} \\ \cline{2-5}
        &  MAE  & RMSE & MAE & RMSE \\ \hline
        W/o Seg. map & 60.1 & 105.0 & 95.6 & 165.4\\
        W/ Seg. map as \cite{shi2019counting}  & 59.5 & 102.2 & 92.3  & 155.3 \\
        W/ Seg. map as SGANet & \bf 58.0 & \bf 100.4 & \bf 89.1  & \bf 150.6 \\
        \hline
    \end{tabular}%
	}
}
\end{table}

\begin{figure*}[t!]
    \centering
    \includegraphics[width=\textwidth]{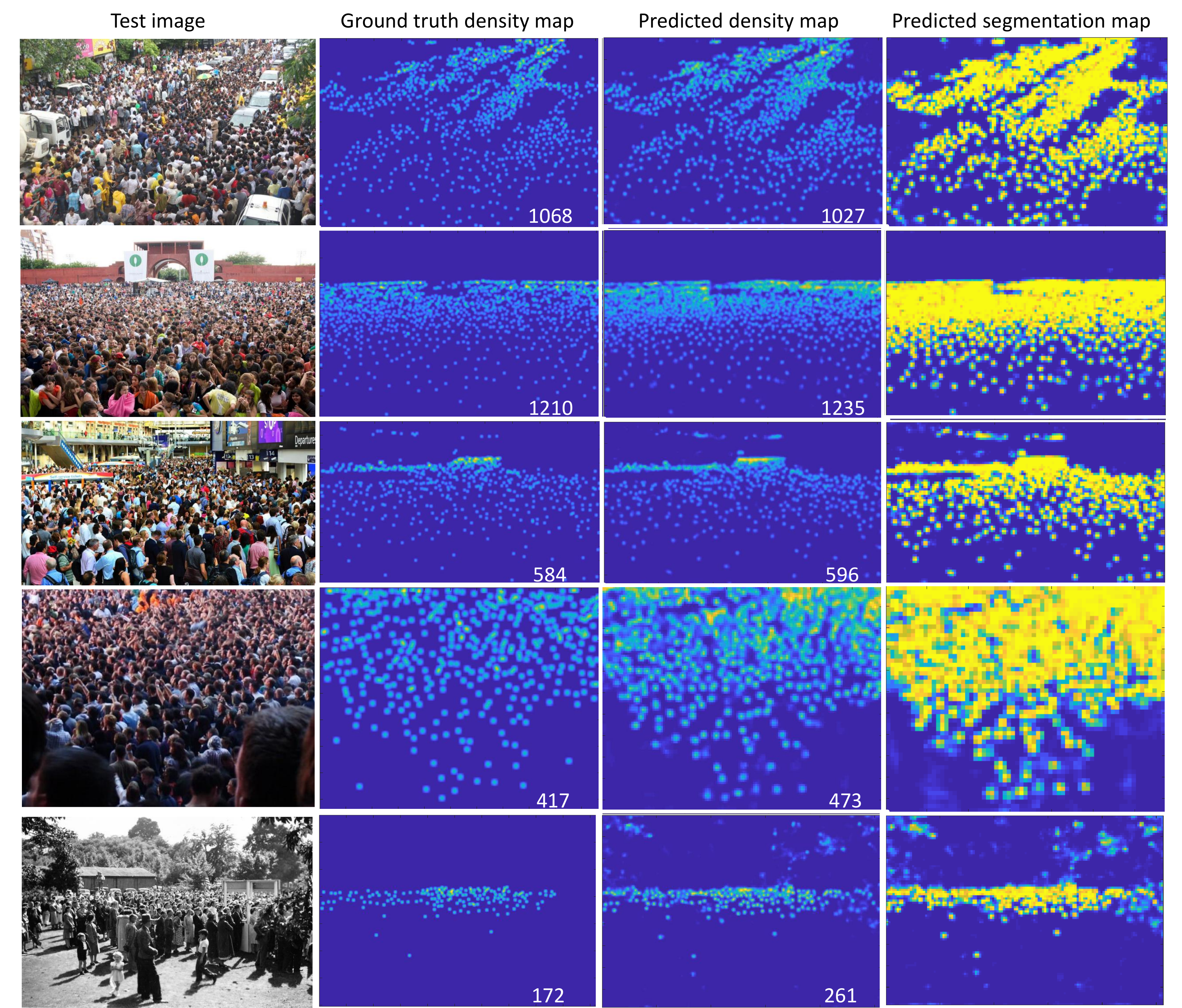}
    \caption{Visualization of estimated density and segmentation maps for five test images from ShanghaiTech part A. The numbers shown on the images in the second and third columns are the ground truth and estimated counts respectively.}
    \label{fig:vis}
\end{figure*}

To give an intuitive evidence on how the attention layer helps for density map estimation, we visualize the estimated attention maps and density maps for five exemplar test images from ShanghaiTech part A. In Figure \ref{fig:vis}, we show the original images, ground truth density maps, predicted density maps and predicted segmentation maps in four columns respectively. The real and predicted counts are also shown on the density maps for a direct comparison. We can see that the prediction errors for the top three examples are relatively low given the accurately predicted segmentation maps. However, the bottom two images suffer from higher errors since the model can not predict accurate foreground regions. For example, the image in the fourth row contains people raising their hands in the air and the hands are easy to be counted since they have similar colours with human faces. In the bottom image, the trees in the background are mistakenly recognised as foreground and result in the over-estimated count.

\section{Discussion and Conclusion} \label{sec:conclusion}
In this paper, we address an important problem crowd counting which can be of great values to intelligent transportation systems. We investigated the effectiveness of \textit{Inception-v3} in crowd counting and proposed a segmentation guided attention network using \textit{Inception-v3} as the backbone. We also proposed a novel curriculum loss function for crowd counting by defining pixel-wise difficulty levels to resolve the issue of scale variance in crowd images. Experimental results on four commonly used datasets demonstrate the proposed SGANet can achieve superior performance due to the combination of \textit{Inception-v3} and the segmentation guided attention layer. The proposed strategy of curriculum learning is also proved to be helpful for a variety of existing crowd counting models in general.

This is the first attempt to use the whole \textit{Inception-v3} model for crowd counting and achieves state-of-the-art performance on commonly used datasets. Although the employed \textit{Inception-v3} model
(with our own modifications) is not designed from scratch,
it is quantitatively shown to be able to achieve superior performance to many specially designed models in the recent
couple of years. To these ends, our work is both disruptive
and important to the crowd counting research community.
Researchers in this community have devoted too much effort to the design of variant CNN architectures and most
of them are based on the pre-trained VGG16 model which
just has insufficient expressive capacity for crowd counting
tasks. In this sense, we believe it is important and necessary to make the community aware of the fact \textit{Inception-v3} is a more suitable architecture for effective crowd counting and divert
the attention of the community to more diverse research directions.

Most existing crowd counting methods including ours in this paper rely on sufficient training data which require extensive data collection and annotation. In real-world applications, it is challenging to get access to sufficient training data for various scenarios (e.g., different camera resolutions, illumination conditions, weather conditions and perspectives). To solve this realistic problem, our future work will focus on weakly supervised learning such as domain adaptation \cite{wang2019unsupervised} and transfer learning \cite{wang2019learning}.
\bibliographystyle{IEEEtran}
\bibliography{ref}

\begin{IEEEbiography}
[{\includegraphics[width=1.2in,height=1.5in,clip,keepaspectratio]{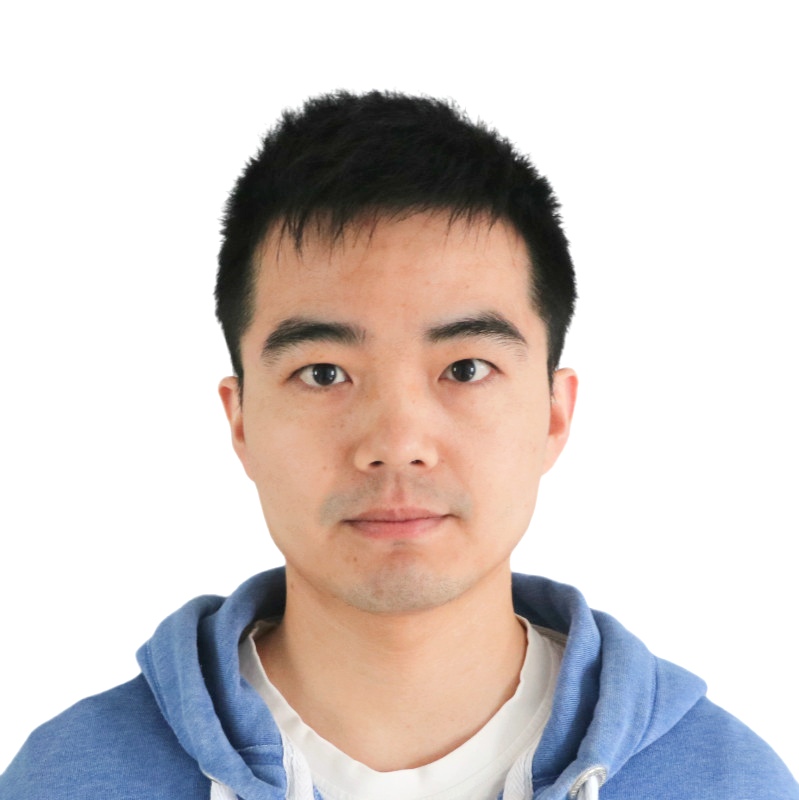}}]{Qian Wang}
is a research associate with department of computer science at Durham University, United Kingdom. His researches focus on deep learning and computer vision. He received his PhD in machine learning from The University of Manchester in 2017, Master's degree in Biomedical Engineering and Bsc in Electronic Engineering in 2013 and 2010 respectively, both from University of Science and Technology of China (Hefei).
\end{IEEEbiography}

\begin{IEEEbiography}
[{\includegraphics[width=1.2in,height=1.5in,clip,keepaspectratio]{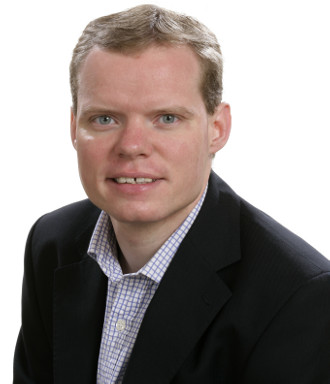}}]{Toby P. Breckon}
is currently a Professor within the Departments of Engineering and Computer Science, Durham University (UK).  His key research interests lie in the domain of computer vision and image processing and he leads a range of research activity in this area.

Prof. Breckon holds a PhD in informatics (computer vision) from the University of Edinburgh (UK).  He has been a visiting member of faculty at the Ecole Supérieure des Technologies Industrielles Avancées (France), Northwestern Polytechnical University (China), Shanghai Jiao Tong University (China) and Waseda University (Japan).

Prof. Breckon is a Chartered Engineer, Chartered Scientist and a Fellow of the British Computer Society. In addition, he is an Accredited Senior Imaging Scientist and Fellow of the Royal Photographic Society. He led the development of image-based automatic threat detection for the 2008 UK MoD Grand Challenge winners [R.J. Mitchell Trophy, (2008), IET Innovation Award (2009)]. His work is recognised as recipient of the Royal Photographic Society Selwyn Award for early-career contribution to imaging science (2011). http://www.durham.ac.uk/toby.breckon/
\end{IEEEbiography}






\end{document}